\documentclass[11pt,a4paper]{article}
\usepackage{authblk}
\usepackage[hyperref]{emnlp2020}
\usepackage{times}
\usepackage{latexsym}
\usepackage{amsfonts}
\usepackage{amssymb}
\usepackage{amsmath}
\usepackage{multirow}
\usepackage{graphicx}
\usepackage{float}
\usepackage{xcolor,colortbl}
\usepackage[normalem]{ulem}
\useunder{\uline}{\ul}{}
\definecolor{RBgrey}{rgb}{0.9,0.9,0.9}
\usepackage{fancyhdr}
\usepackage[ruled,vlined]{algorithm2e}
\include{pythonlisting}
\usepackage{booktabs}
\usepackage{lipsum}

\DeclareMathOperator*{\argmax}{argmax}   
\usepackage{microtype}

\aclfinalcopy 
\def\aclpaperid{3065} 

\title{Data Boost: Text Data Augmentation Through \\ Reinforcement Learning Guided Conditional Generation}

\author[1]{Ruibo Liu}
\author[2]{Guangxuan Xu}
\author[3]{Chenyan Jia}
\author[4]{Weicheng Ma}
\author[5]{Lili Wang}
\author[6]{Soroush Vosoughi}
\affil[1,2,4,5,6] {Department of Computer Science, Dartmouth College}
\affil[3]{Moody College of Communication, University of Texas at Austin}
\affil[1,4,5]{\texttt{\{ruibo.liu.gr,weicheng.ma.gr,lili.wang.gr\}@dartmouth.edu}}
\affil[2]{\texttt{guangxuan.xu.ug@dartmouth.edu}}
\affil[3]{\texttt{chenyanjia@utexas.edu}}
\affil[6]{\texttt{soroush.vosoughi@dartmouth.edu}}

\date{}
\begin{document}
\maketitle
\begin{abstract}
Data augmentation is proven to be effective in many NLU tasks, especially for those suffering from data scarcity. In this paper, we present a powerful and easy to deploy text augmentation framework, \textbf{Data Boost}, which augments data through reinforcement learning guided conditional generation. We evaluate Data Boost on three diverse text classification tasks under five different classifier architectures. The result shows that Data Boost can boost the performance of classifiers especially in low-resource data scenarios. For instance, Data Boost improves F1 for the three tasks by 8.7\% on average when given only 10\% of the whole data for training. We also compare Data Boost with six prior text augmentation methods. Through human evaluations ($N$=178), we confirm that Data Boost augmentation has comparable quality as the original data with respect to readability and class consistency.
\end{abstract}

\section{Introduction}

Data augmentation is a widely-used technique in classification tasks. In the field of computer vision (CV), data is augmented by flipping, cropping, tilting, and altering RGB channels of the original images~\citep{krizhevsky2012imagenet,chatfield2014return,szegedy2015going}; however, similar intuitive and simple strategies do not obtain equal success in NLP tasks. Existing methods tend to produce augmentation with low readability or unsatisfying semantic consistency~\citep{yang2020g}.

\begin{table}[tp!]
\centering
\resizebox{0.48\textwidth}{!}{%
\begin{tabular}{@{}ll@{}}
\toprule
Original                                                                        & So Cute! The baby is very lovely!                                                    \\ \midrule
\begin{tabular}[c]{@{}l@{}}Naive Aug.\\ \textit{Delete} + \textit{Swap}\end{tabular}              & So Cute! is The baby very!                                                           \\ \midrule
\begin{tabular}[c]{@{}l@{}}Word2Vec Aug.\\ \textit{Insert} + \textit{Replace}\end{tabular} & \begin{tabular}[c]{@{}l@{}}So Cute \underline{\textit{adorable}}!\\ The baby is very \underline{\textit{fabulous}}!\end{tabular} \\ \midrule
\begin{tabular}[c]{@{}l@{}}Back Translate Aug.\\ \textit{Eng.} $\rightarrow$ \textit{Fr.} $\rightarrow$ \textit{Eng.}\end{tabular} & \underline{\textit{Cute}}! The baby is very \underline{\textit{cute}}!                                                         \\ \midrule
\textbf{Data Boost}                                                             & \begin{tabular}[c]{@{}l@{}}\textit{\textbf{Look at this adorable baby!}}\\ \textit{\textbf{He is so cute!}}\end{tabular} \\ \bottomrule
\end{tabular}%
}
\caption{A simple demo of existing text data augmentation methods on \textit{positive} sentiment label.}
\vspace{-0.3in}
\label{tab:overview}
\end{table}

Table~\ref{tab:overview} shows some output samples of popular text augmentation methods. Naive methods imitate pixel manipulation in CV, augmenting sentences by adding spelling errors~\citep{xie2017data}, or randomly deleting and swapping tokens~\citep{wei2019eda}. The output of such augmentation methods are often illegible since the word order is disrupted (e.g., \textit{``is The baby very!"}); even worse, crucial feature words (e.g., the word \textit{lovely} which is a signal-carrying word for sentiment detection) could be mistakenly removed through random deletion. A more advanced method is synonym insertion or replacement~\citep{zhang2015character,wang_yang_2015_thats}, which uses Word2Vec~\citep{mikolov2013distributed} to replace words with their synonyms. Such a method respects the original sentence structure but fails to consider the context. It sometimes replaces words with synonyms that are awkward in the full context of the sentence. For example, replacing \textit{lovely} with \textit{fabulous} to get the sentence  \textit{``The baby is fabulous!"}. Recent work leans towards translation-based methods for augmentation~\citep{fadaee2017data,silfverberg2017data}. In particular, \citet{yu2018qanet} proposed a back-translation method that first translates the text to French and then translates back to English, using the noisy output as the augmentation data. Although back-translation is intuitive and valid, its generation skews towards high frequency words (e.g., \textit{cute, lovely} are both back-translated to \textit{cute}), which not only causes repetition but also leads to lexical shrinkage in the augmented data. In a nutshell, existing techniques 
are still far from perfect, partially due to the strong interdependency of syntactic and semantic features in text data. 

In recent years, we have witnessed extensive progress in language models (LM). Large-scale LMs such as BERT~\citep{devlin2019bert}, XLNet~\citep{yang2019xlnet}, and GPT-2~\citep{radford2019language}, are commonly trained on large amounts of text data (e.g., GPT-2 was trained on 8 million web pages that emphasized content diversity). One of the most interesting usages of these models is utilizing them as text generators~\citep{raffel2019exploring,lewis2019bart,dong2019unified}. In this paper, we explore whether we can leverage the generation ability of the state-of-the-art LMs, to \textit{generate} augmented samples for a given target class.

Augmentation samples should exhibit features of the target class. Off-the-shelf LMs cannot be directly used to augment data; since they are not trained for specific contexts, their generation is undirected and random. Conditional LMs can generate text directed by certain condition (e.g., target class), but they require training a LM from scratch with data covering all the conditions. \citet{keskar2019ctrl}, for instance, trained a 1.6 billion-parameter LM conditioned to a variety of control codes. The training is rather costly; however, collecting sufficient data for the training is also tedious, especially in low-resource tasks~\citep{waseem:2016:NLPandCSS}.

We thus present \textbf{Data Boost}: a 
reinforcement learning guided text data augmentation framework built on off-the-shelf LM (GPT-2). Data Boost requires neither collecting extra data nor training a task-specific LM from scratch. We convert GPT-2 into a conditional generator, and for a given task, we guide the generator towards specific class labels during its decoding stage through reinforcement learning. The generated samples can then serve as augmentation data which are similar to the original data in terms of semantics and readability.

The advantages of Data Boost are three-fold: \textit{First}, Data Boost is powerful. We achieve significant advances in three tasks on five different classifiers compared with six related works. \textit{Second}, Data Boost generates sentence-level augmentation. Unlike prior methods that do word-level or phrase-level replacement~\citep{kobayashi2018contextual,wei2019eda}, our augmented data is of much greater variety in terms of vocabulary and sentence structure. Human evaluations also verify the high readability and label consistency of our augmentation. \textit{Third}, Data Boost is easy to deploy. It does not require external datasets or training separate systems (like machine translation model in the back-translation method). Instead, we take the off-the-shelf GPT-2 language model and modify its decoding stage without changing its architecture.

\section{Data Boost}
\label{sec:data_boost}

\begin{figure}[!t]
  \centering
   \includegraphics[width=0.48\textwidth]{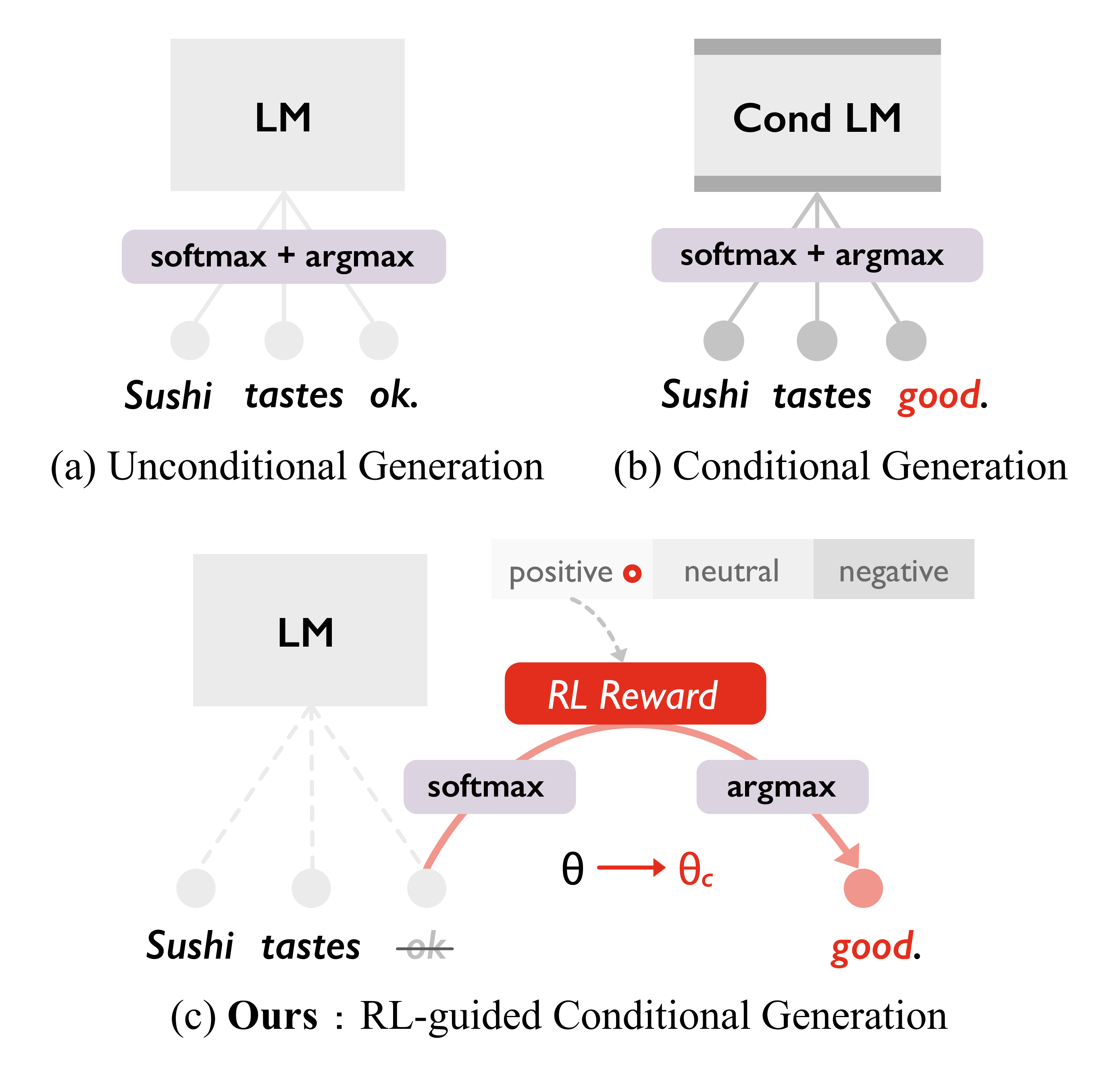}
   
    \caption{General illustration of previous generation models and Data Boost. We add an additional RL stage between the softmax and argmax function, to update the LM hidden-states parameter $\theta$ towards target label (e.g. \textit{positive}).}
    \vspace{-0.2in}
   \label{fig:apr_cond_gen}
 \end{figure}

\subsection{Conditional Generator}
\label{subsec:condgen}

Given tokens $x_{<t} = \{x_0, x_1, ..., x_{t-1}\}$ and accumulated hidden states $h_{<t}^{\theta}$~\footnote{Hidden states are basically key-values pairs in the attention blocks. We denote their values as parameter set $\theta$.} before time step $t$, a vanilla auto-regressive language model (LM) is trained to maximize probability of the next step token $\hat{x}_t$. Normally the model will pick the token that has the highest probability $x_{t}$ as the $t$ step decoding output:

\begin{equation}
    x_t \sim \argmax_{\hat{x}_t} {p(\hat{x}_{t} | x_{<t})} = \textrm{LM}(x_{<t}, h_{<t}^{\theta})
\end{equation}

The generation of such step-by-step decoding is unconditional, since the model is trained on unannotated data (Figure~\ref{fig:apr_cond_gen} (a)). Conditional generation, however, normally needs to train a conditional LM. By modifying the LM architecture to allow for extra input (target label), the conditional LM can model the language and its corresponding label at the same time (Figure~\ref{fig:apr_cond_gen} (b)). Its generation is thus conditional on the label but the training of LM is always costly. 

Different from the above conditional generation method, we keep the architecture of the existing LM unchanged but postpone the argmax function to a later stage. In this way, the output of softmax is still differentiable (as it is a probability over the whole vocab rather than decoded tokens), which allows for gradient-based optimization. As shown in Figure~\ref{fig:apr_cond_gen} (c), we add a reinforcement learning (RL) stage within the gap between the softmax and argmax function. The RL reward (defined in Section~\ref{subsub:rewards}) is where we inject the controlling signal of target label to guide the generation towards the target label. Specifically, in each decoding step, we update the hidden states parameter $\theta$ to the conditional $\theta_c$ in terms of the back-propagated reward after several iterations of RL optimization. The final decoded output shall be conditional on the target label (which is \textit{positive} in Figure~\ref{fig:apr_cond_gen} (c)).

\subsection{Reinforcement Learning Optimization}
\label{subsec:rl}

\subsubsection{Reward}
\label{subsub:rewards}
In the reinforcement learning framework, we define the state at step $t$ as all the generated sequence before $t$ (i.e., $s_t = x_{<t}$), and the action at step $t$ as the $t$-th output token (i.e., $a_t = x_{t}$). The policy $\pi_{\theta}$ is interpreted as the probability we choose token $x_t$ (action $a_t$) given the state $s_t = x_{<t}$, which is the softmax output of the hidden states (i.e., $\pi_{\theta}(a_t|s_t) = \textrm{softmax}(h_{<t}^{\theta})$, and similar for the conditional case).

We define the single-step reward of the conditional generated token $x_t^c$ at step $t$ as:

\begin{equation}
\label{eqa:rewards}
    R(x_t^c) = \mathbb{E}_{t}\left[\frac{\pi_{\theta_{c}}(a_t|s_t)}{\pi_{\theta}(a_t|s_t)}G(x_t^c)\right] - \beta \textrm{KL}(\theta || \theta_{c})
\end{equation}

\noindent where $G(x_t^c)$ is the salience gain that measures how closely the generated token resembles the salient lexicon of the target label, and serves as a guide signal for the conditional generation. We also consider the Kullback–Leibler (KL) divergence between the conditional $\theta_c$ and unconditional distribution of $\theta$ as an auxiliary constraint (with weight $\beta$). Such a reward composition follows the classic PPO (Proximal Policy Optimization)~\citep{schulman2017proximal} form. Note that we are using an off-policy strategy to collect unconditional $(s_t, a_t)$ pairs as trajectory to estimate the conditional reward $R(x_i^c)$. In this way, we are able to perform several iterations of updates on $\theta$ to maximize the reward without changing the sampling policy frequently, which avoids potential instability~\citep{munos2016safe}. As a result, we use the probability ratio between the conditional policy $\pi_{\theta_{c}}$ and the unconditional policy $\pi_{\theta}$ to re-calibrate the reward in the first term of Equation~\ref{eqa:rewards}.

\paragraph{Salience Gain.}\label{para:salience} For a given task that has $K$ classes, we define the salience score of word $x$ belonging to a certain class $c$ as:

\begin{equation}
  \mathcal{S}_{x, c} = \textrm{GM}(\frac{|x \in c|}{\sum\limits_{k=1}^K |x \in c_k|},\  \frac{|x \in c|}{\sum\limits_{x_i \in |V|} |x_i \in c|}) 
\end{equation}

\noindent where $|x \in c|$ refers to the count of word $x$ in samples with class label $c$, $|V|$ is the total vocabulary, and $\textrm{GM}$ is geometric mean of the two terms. The two fractions try to guarantee that both $\mathbb{P}(c|x)$ and $\mathbb{P}(x|c)$ probabilities are high for a word marked as salient. We calculate the salience score for each word and pick the top-N highest words~\footnote{N is a hyperparameter that is related to the size of the dataset. In our case, we set N=500 for sentiment analysis and irony classification tasks, and N=1000 for offense detection task.} as the salient lexicon for class label $c$ (denoted as $w_c$). Compared with other methods such as training a discriminator~\citep{Dathathri2020Plug} or deriving control codes~\citep{keskar2019ctrl}, we find our frequency-based method is relatively simple but efficient especially in data hungry cases, where the performance of a discriminator could be limited given very few training data. 

For the $t$-th step token $x_t^c$ conditional on the target class $c$, we calculate the salience gain as the logarithm summation of cosine similarity with each word in the salient lexicon $w_c$: 

\begin{equation}
  G(x_t^c) = \sum_{w_i \in w_c} \log(\textrm{softmax}(h_{<t}^{\theta_c}) \cdot \textrm{emb}(w_i))
\end{equation}

\noindent We use the embedded vector of $w_i$ and the softmax output of $t$-th step hidden states $h_t^{\theta_c}$ to compute a dot product in the latent space. The salience gain measures how much the current step token resembles the salient lexicon of the target class.

\paragraph{KL Penalty.} It is possible that the conditional policy $\pi_{\theta_{c}}$ drifts away too much from the unconditional policy $\pi_{\theta}$ resulting in an unreadable generation. Therefore, we incorporate a KL divergence penalty term measuring the distance between the two policies, in order to have better insurance that we are optimizing within a trust region. The KL divergence on the policies is computed as:

\begin{equation}
  \textrm{KL}(\theta || \theta_c) = \sum_{i \in [1,t]} \pi_{\theta}(a_i|s_i) \cdot \log \frac{\pi_{\theta}(a_i|s_i)}{\pi_{\theta_c}(a_i|s_i)}
\end{equation}

We deduct KL divergence with weight $\beta$ as a penalty term in the reward function (Equation~\ref{eqa:rewards}). One can either choose a constant $\beta$ or vary it dynamically.

\subsubsection{Policy Gradient}

Given the reward and the definitions described above, we update our policy at $t$-th step as:

\begin{equation}
\label{eqa:policy}
    \theta_c \leftarrow \theta + \eta \sum_{i=1}^k \frac{\triangledown R(x_t^c / T) }{||\triangledown R(x_t^c / T)||}
\end{equation}

\noindent where $\eta$ is the learning rate and $\theta_c$ is the parameter for the conditional hidden states. In general, we follow the classical SGD update rule, but make two main changes: (1) We use temperature parameter $T$ to control the stochastic sampling during token decoding~\cite{keskar2019ctrl}. $T \rightarrow 0$ approximates a greedy decoding strategy that amplifies the peak in the vocab distribution while $T \rightarrow \infty$ makes the distribution more uniform. (2) We sum the normalized gradient of the reward for $k$ steps. $k$ can be treated as the strength of control over the conditional generation. Combining all above definitions, the policy gradient of Data Boost is summarized in Algorithm 1.

\begin{algorithm}[!h]
\SetAlgoLined
\KwIn{Target class label $c$, hidden-states param $\theta$, target KL-divergence $\sigma$.}
 \For{$t=0,1,2,\ldots$}{
  Generate $(a_t|s_t)$ by unconditional policy $\pi_{\theta}$ as trajectories\;
  Estimate reward $R(x_t^c)$ using Eq.~\ref{eqa:rewards}\;
  Compute policy update using Eq.~\ref{eqa:policy} by taking $k$ steps of SGD (via Adam)\;
  \uIf{$\textrm{KL}(\theta||\theta_c) \ge 2\sigma$}{
   $\beta_{t+1} = 2\beta_t$\;
   }
   \ElseIf{$\textrm{KL}(\theta||\theta_c) \le \sigma / 2$}{
   $\beta_{t+1} = \beta_t$ / 2\;
  }
  Return the conditional policy $\pi_{\theta_c}$\;
 }
 \caption{Data Boost Policy Gradient}
\end{algorithm}

We use a dynamic $\beta$ to control the KL penalty within the reward function. The target divergence $\sigma$ depends on the users' need: smaller $\sigma$ means more resemblance to the unconditional generation while larger $\sigma$ provides more space for RL guidance. After several iterations of RL optimization, the updated parameter set $\theta_c$ should be conditional on the target class label, whose feature lexicon contribute to the calculation of the reward $R$. We then use the conditional policy $\pi_{\theta_c}$ (which is based on the hidden states with $\theta_c$) to decode this step token. The token should conform to the specified target class label $c$, since its corresponding hidden states have shifted towards $c$ due to RL optimization.

\section{Tasks \& Datasets}
\label{sec:exp_setup}

We evaluated and compared Data Boost with several state-of-the-art text augmentation methods on the following three tasks:

\noindent \textbf{Offense Detection}\footnote{https://sites.google.com/view/icwsm2020datachallenge} ICWSM 20' Data Challenge dataset ($N = 99,603$) for offensive language detection on tweets. The dataset consists of four classes: \{\textit{normal}, \textit{spam}, \textit{abusive} and \textit{hateful}\} with ratio \{53.9\%, 27.1\%, 14.1\%, 4.9\%\} respectively.

\noindent\textbf{Sentiment Analysis}\footnote{http://alt.qcri.org/semeval2017/task4} SemEval 2017 Task 4A dataset ($N=20,631$) for sentiment analysis in tweets. There are three classes in the dataset: \{\textit{positive}, \textit{neutral} and \textit{negative}\} with ratio \{34.7\%, 49.8\%, 15.5\%\}. 

\noindent\textbf{Irony Classification}\footnote{https://competitions.codalab.org/competitions/17468} SemEval 2018 Task 3A dataset ($N=3,817$) for irony detection in tweets. It has binary classes: \{\textit{ironic}, \textit{non-ironic}\}, with ratio \{50.2\%,  49.8\%\}.

Offense Detection and Irony Classification are popular NLU tasks that are low-resource. Sentiment Analysis, though seemingly well-resolved according to some literature~\citep{baziotis2017datastories,cliche2017bb_twtr}, is reported to have severe overfitting problems when given extremely limited training data~\citep{elming2014robust,severyn2015twitter}. We choose challenging datasets varying in total data size ($N \approx 80\textrm{k}, 17\textrm{k}, 3\textrm{k}$) and the number of class ($\#\ \textrm{of class} = 4, 3, 2$) for a realistic evaluation of our framework. 

\begin{figure*}[h!t]
  \centering
   \includegraphics[width=\textwidth]{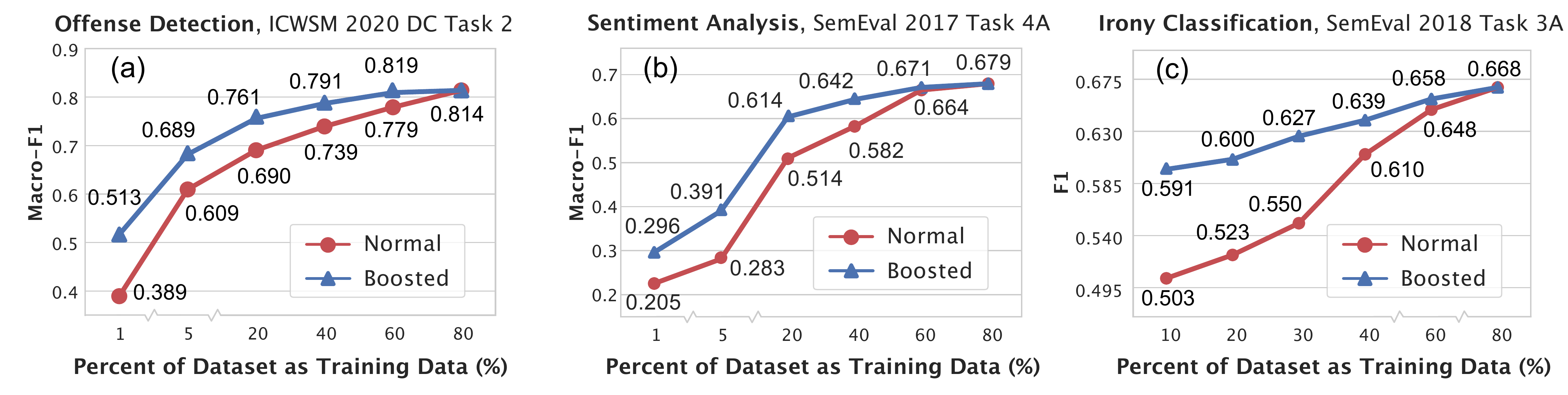}
    \caption{Performance with and without Data Boost on three classification tasks: (a) Offensive Language Detection on Tweets, (b) Sentiment Analysis in Twitter, (c) Irony Detection in English Tweets. The performance is reported by the Macro-F1 (F1 for binary task) of a BERT classifier averaged on five times repeated experiments.} 
   \label{fig:exp1_gain}
 \end{figure*}

We removed all punctuation, stop words, hashtags and url links in the samples for all datasets. Samples whose length was above 30 tokens were filtered out (around 2\% of the data on average) as 30 was also used as the max sequence length for Data Boost generation. We further split the data into training and test set by the ratio \{80\%, 20\%\}, and maintained the original class distributions. We made sure the distributions remained the same in all of our experiments.

\section{Experiments\footnote{We run our generation and classification training on 2 RTX 2080 GPUs for all the experiments. The average time for Data Boost to generate a 30 token long sequence is under 1 second.}}
\label{sec:evaluation}

We conducted extensive experiments to answer the following three overarching questions about Data Boost:

\subsection{Does Data Boost Improve Performance?}
\label{subsec:eva_performance}

Several sets of data starvation tests are prepared, each using restricted fractions of the total data as training data. We keep test data the same (20\% of the whole dataset) but gradually decrease the size of training data from 80\% (as the fully-loaded case) to 1\% (as the extremely low-resource case). We run both normal training and boosted training over the following training set fractions (\%): \{1\%, 5\%, 20\%, 40\%, 60\%, 80\%\} of the total data for both Offense Detection and Sentiment Analysis. Since the dataset for Irony Classification is small ($N = 3,810$), we use the following fractions: \{10\%, 20\%, 30\%, 40\%, 60\%, 80\%\}. Note that for boosted training we add augmentation samples to training data until the training data size reaches 80\% of the total size (same as fully-loaded size), to make sure that the size of the training set does not influence the results.

Figure~\ref{fig:exp1_gain} shows the performance of the BERT~\citep{devlin2019bert} (bert-base-cased) classifier fine-tuned on the three tasks with and without Data Boost over all training set fractions. Data Boost has greater improvements on extremely low-resource cases: we achieve absolute F1 increases of 12.4\% (Offense), 9.1\% (Sentiment) and 8.8\% (Irony) when using only 1\% (10\% for the Irony task) of the original data as training data. The results show that Data Boost can benefit a wide range of tasks with different characteristics. Also, since we used BERT as our classifier, which is already pre-trained on a large corpus, our results confirm that Data Boost can even improve the performance of large-scale LM based classifiers.

\begingroup
\setlength{\tabcolsep}{4pt}
\begin{table*}[!t]
\centering
\resizebox{0.98\textwidth}{!}{%
\begin{tabular}{@{}cccccccccc@{}}
\toprule
\multirow{2}{*}{\textbf{Classifier}} & \multicolumn{3}{c}{\textbf{Offense Detection}} & \multicolumn{3}{c}{\textbf{Sentiment Analysis}} & \multicolumn{3}{c}{\textbf{Irony Classification}} \\ \cmidrule(l){2-10} 
 & 20\% ($\times$2) & 40\% ($\times$2) & 80\% & 20\% ($\times$2) & 40\% ($\times$2) & 80\% & 20\% ($\times$2) & 40\% ($\times$2) & 80\% \\ \midrule
\textbf{CNN~\citeyearpar{kim2014convolutional}} & 0.668 & 0.744 & 0.785 & 0.458 & 0.502 & 0.557 & 0.573 & 0.589 & 0.609 \\
\textit{\textbf{+ Data Boost}} & 0.711 & 0.767 & - & 0.477 & 0.527 & - & 0.585 & 0.598 & - \\ \midrule
\textbf{Bi-LSTM + Attn~\citeyearpar{zhou2016attention}} & 0.696 & 0.744 & 0.788 & 0.439 & 0.515 & 0.564 & 0.468 & 0.554 & 0.598 \\
\textit{\textbf{+ Data Boost}} & \textbf{0.764} & 0.778 & - & 0.513 & 0.542 & - & 0.550 & 0.579 & - \\ \midrule
\textbf{Transformer~\citeyearpar{vaswani2017attention}} & 0.693 & 0.754 & 0.794 & 0.371 & 0.458 & 0.551 & 0.556 & 0.561 & 0.601 \\
\textit{\textbf{+ Data Boost}} & 0.740 & 0.781 & - & 0.502 & 0.521 & - & 0.577 & 0.593 & - \\ \midrule
\textbf{BERT~\citeyearpar{devlin2019bert}} & 0.716 & 0.757 & 0.814 & 0.514 & 0.582 & 0.679 & 0.523 & 0.610 & \textbf{0.668} \\
\textit{\textbf{+ Data Boost}} & 0.720 & \textbf{0.784} & - & 0.610 & 0.642 & - & 0.596 & 0.639 & - \\ \midrule
\textbf{XLNet~\citeyearpar{yang2019xlnet}} & 0.680 & 0.718 & \textbf{0.834} & 0.624 & 0.643 & \textbf{0.697} & 0.632 & 0.639 & 0.664 \\
\textit{\textbf{+ Data Boost}} & 0.693 & 0.755 & - & \textbf{0.636} & \textbf{0.657} & - & \textbf{0.642} & \textbf{0.662} & - \\ \bottomrule
\end{tabular}%
}
\caption{The classifier-agnostic experiments for five main-stream classifiers. We show the results before and after we apply Data Boost on two settings of training data: 20\% original + 20\% boosting data, and 40\% original + 40\% boosting data. We also list the performance of 80\% as training data (full) as reference.}
\label{tab:classifier_agnostic}
\end{table*}
\endgroup

\begin{figure}[h]
  \centering
   \includegraphics[width=0.5\textwidth]{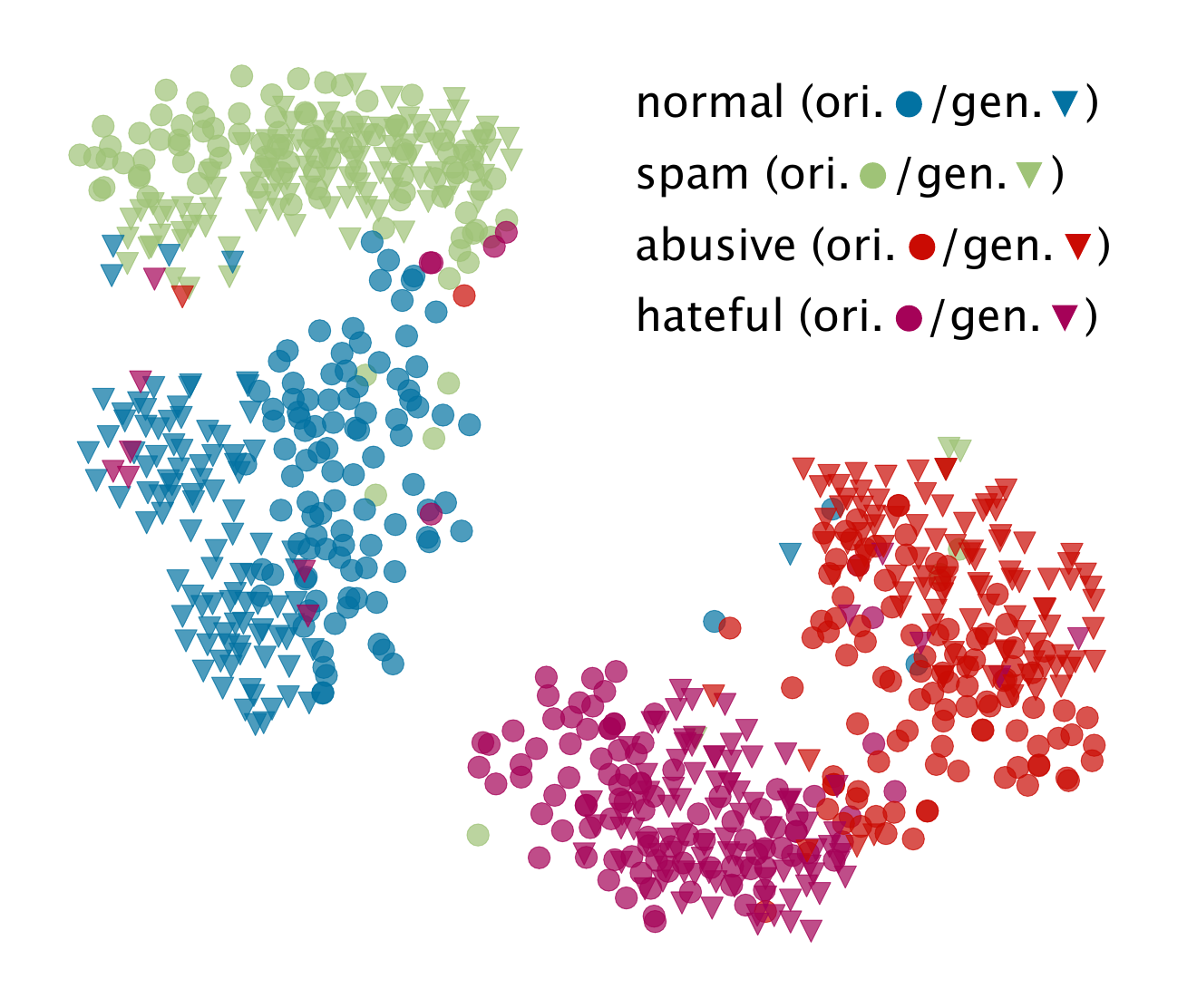}
    \caption{t-SNE visualization of the vectorized original and Data Boost augmented sentences in the offense detection task. The augmented sentences (triangles) mostly overlap with the original sentences (circles), suggesting that augmented sentences maintain the original class distribution.} 
   \label{fig:exp2_tsne}
\end{figure}

\subsection{Does Boosted Data Resemble the Original?}
\label{subsec:eva_label_holding}

\begin{table}[h]
\centering
\resizebox{0.48\textwidth}{!}{%
\begin{tabular}{@{}ccccc@{}}
\toprule
\multicolumn{5}{c}{\textbf{Offense Detection}} \\ \midrule
\textit{ratio} & 100\% / 0 & 75\% / 25\% & 50\% / 50\% & 25\% / 75\% \\ \midrule
\textbf{F1} & 0.814 & $\downarrow$0.005 & $\downarrow$0.028 & $\downarrow$0.060 \\
\textbf{PPL} & 12.28 & 14.03 & 17.71 & 21.53 \\ \midrule
\multicolumn{5}{c}{\textbf{Sentiment  Analysis}} \\ \midrule
\textit{ratio} & 100\% / 0 & 75\% / 25\% & 50\% / 50\% & 25\% / 75\% \\ \midrule
\textbf{F1} & 0.679 & $\downarrow$0.008 & $\downarrow$0.037 & $\downarrow$0.065 \\
\textbf{PPL} & 22.52 & 25.76 & 29.41 & 36.01 \\ \midrule
\multicolumn{5}{c}{\textbf{Irony Classification}} \\ \midrule
\textit{ratio} & 100\% / 0 & 75\% / 25\% & 50\% / 50\% & 25\% / 75\% \\ \midrule
\textbf{F1} & 0.668 & $\downarrow$0.010 & $\downarrow$0.029 & $\downarrow$0.068 \\
\textbf{PPL} & 33.47 & 36.85 & 40.71 & 45.53 \\ \bottomrule
\end{tabular}%
}
\caption{Evaluation of the generation quality in terms of F1 deterioration and perplexity (PPL) increase. We keep the training data size the same, but control the ratio of original/boosting. The first results column corresponds to no boosting.}
\label{tab:functional_same}
\end{table}

A common concern in text data augmentation is whether the augmented sentences preserve the quality of the original data. This is especially true for generation-based methods since we create new sentences rather than simply replace tokens to produce augmented data. We will illustrate the quality of our data generation with two approaches: (1) Visualizing the class distribution of the original and the augmented data (2) By using the boosting ratio experiments described in Section \ref{subsec:eva_performance} to see whether data augmentation causes performance deterioration and perplexity increase.

\begingroup
\setlength{\tabcolsep}{3pt}
\begin{table*}[!t]
\centering
\resizebox{\textwidth}{!}{%
\begin{tabular}{@{}cccccccccc@{}}
\toprule
\multirow{2}{*}{\textbf{Methods}} & \multicolumn{3}{c}{\textbf{Offense Detection}} & \multicolumn{3}{c}{\textbf{Sentiment Analysis}} & \multicolumn{3}{c}{\textbf{Irony Classification}} \\ \cmidrule(l){2-10} 
 & 10\% & 40\% & PPL & \multicolumn{1}{l}{10\%} & 40\% & PPL & \multicolumn{1}{l}{10\%} & 40\% & PPL \\ \midrule
\begin{tabular}[c]{@{}c@{}}\textbf{Naive Aug.}~\citep{coulombe2018text}\\ (keyboard / OCR / spelling error)\end{tabular} & 0.670 & 0.661 & 209.17 & 0.566 & 0.609 & 318.99 & 0.567 & 0.532 & 195.81 \\ \midrule
\begin{tabular}[c]{@{}c@{}}\textbf{Word Replace Aug.}~\citep{niu2018adversarial}\\ (synonyms + antonym from WordNet)\end{tabular} & 0.675 & 0.663 & 41.43 & 0.585 & 0.606 & 63.17 & 0.511 & 0.572 & \textbf{57.52} \\ \midrule
\begin{tabular}[c]{@{}c@{}}\textbf{EDA}~\citep{wei2019eda}\\ (randomly delete, swap, etc.)\end{tabular} & 0.637 & 0.629 & 37.37 & 0.560 & 0.608 & \textbf{41.22} & 0.530 & 0.515 & 76.07 \\ \midrule
\begin{tabular}[c]{@{}c@{}}\textbf{Word2Vec Aug.}~\citep{wang_yang_2015_thats}\\ (insert, replace using Word2Vec)\end{tabular} & 0.673 & 0.720 & 376.43 & 0.557 & 0.619 & 561.31 & 0.548 & 0.585 & 384.61 \\ \midrule
\begin{tabular}[c]{@{}c@{}}\textbf{Contextual Word Embs Aug.}~\citep{kobayashi2018contextual}\\ (insert, replace using Bi-RNN LM)\end{tabular} & 0.663 & 0.713 & 1729.62 & 0.610 & 0.627 & 1043.18 & 0.518 & 0.593 & 1146.40 \\ \midrule
\begin{tabular}[c]{@{}c@{}}\textbf{Back-Translation Aug.}~\citep{yu2018qanet}\\ (Eng. $\rightarrow$ Fr. $\rightarrow$ Eng. as aug. text)\end{tabular} & 0.655 & 0.724 & 345.23 & \textbf{0.617} & 0.620 & 474.29 & 0.520 & 0.541 & 423.32 \\ \midrule
\begin{tabular}[c]{@{}c@{}}\textbf{Ours: Data Boost}\\ (RL-guided conditional generation)\end{tabular} & \textbf{0.695} & \textbf{0.784} & \textbf{35.18} & 0.591 & \textbf{0.642} & 56.23 & \textbf{0.591} & \textbf{0.639} & 77.40 \\ \bottomrule
\end{tabular}%
}
\caption{Performance comparison with other text augmentation methods. 10\%: 10\% original data + 30\% augmented data; 40\%: 40\% original data + 40\% augmented data. We report the F1 score of the BERT classifier over five times repeat experiments. We also report the perplexity score (PPL) of the augmented data (10,000 randomly sampled) from different methods scored by kenLM language models trained on the training data of each task.}
\label{tab:literature}
\end{table*}
\endgroup

For visualization, we randomly pick 400 (100 for each class) original and generated sentences in the Offense Detection task (since it has the largest number of classes) and vectorize with Sentence-BERT~\citep{reimers2019sentence}. We apply t-SNE~\citep{maaten2008visualizing} to these vectors and plot their 2-D representations in Figure~\ref{fig:exp2_tsne}. From the figure, we can see that our RL-based algorithm manages to guide the generation towards the target labels, and for the most part, the distribution of generated sentences matches that of the original data.

Ratio-controlled experiments test the quality of boosted data by comparing training performance. If training on augmented dataset has comparable performance (F1) as training on purely original data, one may infer that the quality of the augmentation data resembles that of the original data. We also use perplexity (PPL) as an auxiliary metric to evaluate the augmentation quality. We trained three language models using kenLM~\citep{heafield2011kenlm} on the original data of the three tasks. We use these models to calculate the perplexity of the ratio-controlled sets.

In Table~\ref{tab:functional_same} we show the F1 deterioration and perplexity increase (higher perplexity means poorer fit to the LM) for different augmentation ratios. Even when we use 25\% original data fused with 75\% generated samples, the F1 score only undergoes a slight decrease (0.06, absolute) compared to when using 100\% original data. We found that the perplexity also did not substantially increase even with higher boosting ratios.

\subsection{Is Data Boost Classifier-Agnostic?}
\label{subsec:eva_classifier_agnostic}
We have shown Data Boost to be effective when used in conjunction with a BERT classifier, but can the performance be replicated with other classifiers? In other words, is Data Boost a classifier-agnostic augmentation method? To answer this question, we ran experiments on four other main-stream classifiers, including the plain CNN classifier~\citep{kim2014convolutional}, the Bi-LSTM with attention mechanism~\citep{zhou2016attention}, the self-attention based Transformer network~\citep{vaswani2017attention}, and another LM-based classifier XLNet~\citep{yang2019xlnet} for comparison. We trained all classifiers on three different training data settings: \{20\%, 40\%, 80\%\} of the total data used as training data, the first two datasets are doubled in size using Data Boost augmentation. As shown in Table \ref{tab:classifier_agnostic}, Data Boost generally improves the performance of all the classifiers (from 1\% to 13\%, absolute), regardless of the classifier architecture. Moreover, we find Data Boost is not only effective for relatively simple classifiers (e.g., CNN), but also beneficial to complex LM-based classifiers (e.g., BERT and XLNet), which are already trained on a large corpus and generally used as very strong baselines for text classification tasks.

\begin{table*}[]
\centering
\resizebox{\textwidth}{!}{%
\begin{tabular}{@{}llll@{}}
\toprule
\textbf{Class} &
  \multicolumn{3}{l}{\textbf{Generation Samples}} \\ \midrule
\textit{ironic} &
  \multicolumn{3}{l}{\begin{tabular}[c]{@{}l@{}}\underline{freezing} cold winter air can be a real treat but if your room temperature stays below \underline{freezing} \\ for a long time then the \underline{best} way to cool down is death.\end{tabular}} \\ \cmidrule(r){1-1}
\textit{non-ironic} &
  \multicolumn{3}{l}{\begin{tabular}[c]{@{}l@{}}\underline{FoxNewscom} reporter Michelle Fields is being sued by a Republican donor for disrespecting \\ him and his family the Republican National Committee announced Wednesday.\end{tabular}} \\ \midrule
\textit{positive} &
  \multicolumn{3}{l}{\begin{tabular}[c]{@{}l@{}}\underline{Congratulations} to our friends at Bored Panda Pizza for the \underline{wonderful} promotion that they have\\  done We are very \underline{happy} and \underline{proud} to be able to share.\end{tabular}} \\ \cmidrule(r){1-1}
\textit{neutral} &
  \multicolumn{3}{l}{\begin{tabular}[c]{@{}l@{}}\underline{Results} of the study revealed that the amount of protein ingested was similar in each group but \\ not significantly different in total fat total carbohydrate or total protein.\end{tabular}} \\ \cmidrule(r){1-1}
\textit{negative} &
  \multicolumn{3}{l}{\begin{tabular}[c]{@{}l@{}}\underline{disappointed} by the news media reports In the United States media have been covering reports \\ on the \underline{killing} of a woman by two men on a train.\end{tabular}} \\ \midrule
\textit{normal} &
  \multicolumn{3}{l}{\begin{tabular}[c]{@{}l@{}}\underline{Im} not a doctor or any other medical profession Im just trying to make this post useful to others \\ who are looking through this topic.\end{tabular}} \\ \cmidrule(r){1-1}
\textit{spam} &
  \multicolumn{3}{l}{\underline{Black Friday} sales on \underline{Xbox One} begin today Nov at am ET Heres everything you can find in Black.} \\ \cmidrule(r){1-1}
\textit{abusive} &
  \multicolumn{3}{l}{\begin{tabular}[c]{@{}l@{}}\underline{sick} of all the crap If youve been following the news you know that the \underline{Trump} administration \\ and Democrats have been \underline{attacking} President \underline{Trump} executive.\end{tabular}} \\ \cmidrule(r){1-1}
\textit{hateful} &
  \multicolumn{3}{l}{\begin{tabular}[c]{@{}l@{}}\underline{idiot} how does she know that you are \underline{fucking} with her I dont want to see a \underline{stupid} person like you \\ get \underline{raped} by any \underline{fucking} person.\end{tabular}} \\ \bottomrule
\end{tabular}%
}
\caption{Sample generation of Data Boost for all classes from three tasks. Salient words are \underline{underlined}.}
\label{tab:sample_generation}
\end{table*}

\vspace{2pt}
\noindent Table~\ref{tab:sample_generation} shows sample generations by Data Boost.

\section{Comparison with Related Work} 
\label{sec:related_work}

Table~\ref{tab:literature} compares the performance of Data Boost with six prior text augmentation methods on all three tasks and using a BERT classifier. Naive methods~\citep{coulombe2018text,xie2017data} and translation-based methods~\citep{fadaee2017data,sennrich2016improving} treated data noise either from artificial typos or translation errors as augmentation. \citet{wei2019eda} proposed EDA which is a combination of token-level augmentation (randomly delete, swap, etc.); they reported modest improvement (0.8\% on average) on several benchmark datasets. \citet{zhang2015character} performed character-level augmentation. These methods were usually compromised by low readability and flawed syntactic structure. Other methods utilized external resources to improve augmentation quality. 
For example, \citet{wang_yang_2015_thats} leveraged Word2Vec to extract synonyms. \citet{kobayashi2018contextual} trained a Bi-RNN LM to propose replacements that are context-aware. Our tests find that these methods have higher perplexity than others. The reason could be that Word2Vec does not take context into account, while LM replacement highly depends on the quality of self-trained LM. Data Boost, however, is built on a state-of-the-art LM (GPT-2) and generates augmentations from scratch using RL rather than by replacement. Data Boost outperforms the other methods in the majority of the experiments (Table~\ref{tab:literature}).

A few words about conditional generation: CTRL~\citep{keskar2019ctrl} and BART~\citep{lewis2019bart} are large-scale conditional LMs trained on self-collected data. PPLM~\citep{Dathathri2020Plug} does conditional generation through perturbing the vocabulary distribution during token decoding. These methods have not been explored for text augmentation applications. More importantly, they do not use reinforcement learning to have fine-grained control over generation, which we found especially helpful when dealing with multiple labels within the same task.

\section{Human Evaluation}
\label{sec:human_judgement}

\subsection{Experimental Design}

We conducted human evaluation on Amazon Mechanical Turk (MTurk) in May 2020. Participants ($N$ = 178) were randomly assigned to evaluate one of the three tasks, respectively Irony Classification ($n$ = 60), Sentiment Analysis ($n$ = 58), and Offense Detection ($n$ = 60). Participants were all from the United States and above 18 years old. The average age of participants was 36.92 years-old. More than half (57.3\%) of participants were male, and 42.1\% were female, one participant self-report gender as other. Each participant was paid 75 cents for their participation in this study.

\begin{table*}[t!]
\centering
\resizebox{0.85\textwidth}{!}{%
\begin{tabular}{@{}llcccccc@{}}
\toprule
\multirow{2}{*}{\textbf{Task}} &
  \multirow{2}{*}{\textbf{Class}} &
  \multicolumn{3}{c}{\textbf{Mean (SD)}} &
  \multirow{2}{*}{\textit{F}} &
  \multirow{2}{*}{\textit{df}} &
  \multicolumn{1}{l}{\multirow{2}{*}{$p$-value}} \\
 &
   &
  Original &
  Data Boost &
  Vanilla &
   &
   &
  \multicolumn{1}{l}{} \\ \midrule
\multirow{2}{*}{\begin{tabular}[c]{@{}l@{}}Irony\\ Classification\end{tabular}} &
  \textit{ironic} &
  4.65 (1.27) &
  5.04 (1.16) &
  5.04 (1.10) &
  2.21 &
  177 &
  0.11 \\
 &
  \textit{non-ironic} &
  4.76 (1.31) &
  4.96 (1.17) &
  4.74 (1.36) &
  0.53 &
  177 &
  0.59 \\ \cmidrule(r){1-2}
\multirow{3}{*}{\begin{tabular}[c]{@{}l@{}}Sentiment\\ Analysis\end{tabular}} &
  \textit{positive} &
  4.92 (1.10) &
  4.90 (1.17) &
  4.51 (1.38) &
  2.03 &
  171 &
  0.13 \\
 &
  \textit{neutral} &
  4.26 (1.39) &
  4.83 (1.23) &
  4.88 (1.16) &
  4.40 &
  171 &
  0.01** \\
 &
  \textit{negative} &
  4.41 (1.46) &
  4.72 (1.19) &
  4.94 (1.06) &
  2.61 &
  171 &
  0.08 \\ \cmidrule(r){1-2}
\multirow{4}{*}{\begin{tabular}[c]{@{}l@{}}Offense\\ Detection\end{tabular}} &
  \textit{hateful} &
  4.35 (1.51) &
  4.53 (1.43) &
  5.00 (1.27) &
  3.42 &
  177 &
  0.04* \\
 &
  \textit{abusive} &
  4.42 (1.38) &
  4.61 (1.21) &
  5.05 (1.19) &
  3.82 &
  177 &
  0.02* \\
 &
  \textit{spam} &
  4.43 (1.60) &
  4.86 (1.17) &
  4.63 (1.40) &
  1.39 &
  177 &
  0.25 \\
 &
  \textit{normal} &
  4.83 (1.25) &
  5.15 (1.04) &
  4.92 (1.27) &
  1.11 &
  177 &
  0.33 \\ \bottomrule
\end{tabular}%
}
\caption{Human evaluation results on readability. $p$-value describes the significance of difference. (* corresponds to $p<0.05$, ** to $p<0.01$
and *** to $p<0.001$.)}
\label{tab:human_readability}
\end{table*}

\subsection{Procedures}

For each class, participants were asked to read three samples from each version (the original, unconditionally generated (vanilla GPT-2), and RL-conditional generated(Data Boost)). They were not informed of actual labels and versions of samples. After reading, participants were shown the actual label and version of those samples they just read. They were then asked to answer a series of questions about label agreement (e.g., \textit{``How much do you agree with the assigned class label?"} on a 7-point scale (1-strongly disagree to 7-strongly agree)). Additionally, they were asked to rate the readability of samples on a 7-point scale (lower scores correspond to lower readability and vice versa). The readability measure included five items adapted from previous studies~\citep{graefe2018readers}, namely well-written, concise, comprehensive, coherent, and clear.

\subsection{Results}
\subsubsection{Label Agreement}
We conducted paired sample t-tests to examine how much participants agreed with the assigned labels. To conduct an ablation study, we included samples generated using vanilla GPT-2 and Data Boost. Compared to the vanilla GPT-2, Data Boost samples received higher label agreement scores in eight out of nine classes. Five of which were statistically significantly ($p<$ .05) higher. No statistically significant differences were seen between the original and boosted data, except for the \textit{spam} and \textit{normal} class in Offense Detection ($p=$ .02 and $p=$ .03). This result further confirms that Data Boost samples look very similar to the original samples and that Data Boost generates higher quality samples than the vanilla GPT-2. 

\subsubsection{Readability}
We conducted several one-way analyses of variance (ANOVA) to test whether there were any statistically significant differences in the readability of the three models (Table~\ref{tab:human_readability}). 
There were no significant differences for six of the classes. Curiously, for the \textit{neutral} (Sentiment), \textit{abusive} (Offense) and \textit{hateful} (Offense) labels, both Data Boost and vanilla GPT-2 generated samples were rated as more readable than the original samples ($p<$ .05). This could be explained by the fact that original samples are generally noisy tweets. These results indicate that the Data Boost generation has similar readability as the vanilla GPT-2 or original samples.

\section{Limitations}
\label{sec:limitation}

In this section we discuss the limitations of Data Boost. 
The performance gain achieved by using Data Boost could be marginal on certain tasks, especially those whose classes cannot be modeled well by lexical features. For example, we experimented with Data Boost for metaphor detection using the LCC dataset~\citep{mohler2016introducing}, sarcasm classification using the GHOSH dataset~\citep{ghosh2017magnets}, and formality detection using the GYAFC formality style transfer dataset~\citep{rao2018dear}. We saw marginal improvements in the tasks, with an absolute increase in F1 scores of 1.3\%, 0.9\%, and 0.7\% for the three tasks respectively (in the extreme data scarcity case, where we expect Data Boost to help the most; i.e., when boosting 1\% of the original data to 80\%). We found that it was difficult for our model to extract explicit lexical features for the \textit{metaphor}, \textit{sarcastic}, and \textit{formal} classes. This could be because syntactic features play a role in these classes. It is challenging for Data Boost to compose meaningful augmentation in such cases, given that our guidance on the generation is token-by-token.

\section{Conclusion}
\label{sec:conclusion}

We have proposed a powerful and easy to deploy approach to augment text data through conditional generation. By leveraging an off-the-shelf language model (GPT-2), we successfully guide the generation towards a specified direction (i.e, target class), with the help of reinforcement learning.
We find that Data Boost improves the performance of classification tasks, is classifier-agnostic, and that it surpasses several prior augmentation methods in three diverse classification tasks. 

In the future, we plan to implement a more sophisticated guidance for the augmentation by adding syntactic and position features to the reward function, to enable augmentation of more diverse types of text data. The code will be made available upon request.

\section*{Acknowledgement} 
We sincerely thank the reviewers for their insightful comments and suggestions that helped improve the paper. This research was supported in part by the Dartmouth Burke Research Initiation Award and the Amazon Research Award. 

\bibliographystyle{acl_natbib}
\bibliography{anthology,emnlp2020}

\begin{thebibliography}{41}
\expandafter\ifx\csname natexlab\endcsname\relax\def\natexlab#1{#1}\fi

\bibitem[{Baziotis et~al.(2017)Baziotis, Pelekis, and
  Doulkeridis}]{baziotis2017datastories}
Christos Baziotis, Nikos Pelekis, and Christos Doulkeridis. 2017.
\newblock \href {https://www.aclweb.org/anthology/S17-2126.pdf} {Datastories at
  semeval-2017 task 4: Deep lstm with attention for message-level and
  topic-based sentiment analysis}.
\newblock In \emph{Proceedings of the 11th international workshop on semantic
  evaluation (SemEval-2017)}, pages 747--754.

\bibitem[{Chatfield et~al.(2014)Chatfield, Simonyan, Vedaldi, and
  Zisserman}]{chatfield2014return}
Ken Chatfield, Karen Simonyan, Andrea Vedaldi, and Andrew Zisserman. 2014.
\newblock \href {https://arxiv.org/pdf/1405.3531.pdf} {Return of the devil in
  the details: Delving deep into convolutional nets}.
\newblock \emph{arXiv preprint arXiv:1405.3531}.

\bibitem[{Cliche(2017)}]{cliche2017bb_twtr}
Mathieu Cliche. 2017.
\newblock \href {https://www.aclweb.org/anthology/S17-2094.pdf} {Bb\_twtr at
  semeval-2017 task 4: Twitter sentiment analysis with cnns and lstms}.
\newblock In \emph{Proceedings of the 11th International Workshop on Semantic
  Evaluation (SemEval-2017)}, pages 573--580.

\bibitem[{Coulombe(2018)}]{coulombe2018text}
Claude Coulombe. 2018.
\newblock \href {https://arxiv.org/pdf/1812.04718.pdf} {Text data augmentation
  made simple by leveraging nlp cloud apis}.
\newblock \emph{arXiv preprint arXiv:1812.04718}.

\bibitem[{Dathathri et~al.(2020)Dathathri, Madotto, Lan, Hung, Frank, Molino,
  Yosinski, and Liu}]{Dathathri2020Plug}
Sumanth Dathathri, Andrea Madotto, Janice Lan, Jane Hung, Eric Frank, Piero
  Molino, Jason Yosinski, and Rosanne Liu. 2020.
\newblock \href {https://openreview.net/forum?id=H1edEyBKDS} {Plug and play
  language models: A simple approach to controlled text generation}.
\newblock In \emph{International Conference on Learning Representations}.

\bibitem[{Devlin et~al.(2019)Devlin, Chang, Lee, and
  Toutanova}]{devlin2019bert}
Jacob Devlin, Ming-Wei Chang, Kenton Lee, and Kristina Toutanova. 2019.
\newblock \href {https://arxiv.org/pdf/1810.04805.pdf} {Bert: Pre-training of
  deep bidirectional transformers for language understanding}.
\newblock In \emph{Proceedings of the 2019 Conference of the North American
  Chapter of the Association for Computational Linguistics: Human Language
  Technologies, Volume 1 (Long and Short Papers)}, pages 4171--4186.

\bibitem[{Dong et~al.(2019)Dong, Yang, Wang, Wei, Liu, Wang, Gao, Zhou, and
  Hon}]{dong2019unified}
Li~Dong, Nan Yang, Wenhui Wang, Furu Wei, Xiaodong Liu, Yu~Wang, Jianfeng Gao,
  Ming Zhou, and Hsiao-Wuen Hon. 2019.
\newblock \href {https://arxiv.org/pdf/1905.03197.pdf} {Unified language model
  pre-training for natural language understanding and generation}.
\newblock In \emph{Advances in Neural Information Processing Systems}, pages
  13042--13054.

\bibitem[{Elming et~al.(2014)Elming, Plank, and Hovy}]{elming2014robust}
Jakob Elming, Barbara Plank, and Dirk Hovy. 2014.
\newblock Robust cross-domain sentiment analysis for low-resource languages.
\newblock In \emph{Proceedings of the 5th Workshop on Computational Approaches
  to Subjectivity, Sentiment and Social Media Analysis}, pages 2--7.

\bibitem[{Fadaee et~al.(2017)Fadaee, Bisazza, and Monz}]{fadaee2017data}
Marzieh Fadaee, Arianna Bisazza, and Christof Monz. 2017.
\newblock \href {https://www.aclweb.org/anthology/P17-2090.pdf} {Data
  augmentation for low-resource neural machine translation}.
\newblock In \emph{Proceedings of the 55th Annual Meeting of the Association
  for Computational Linguistics (Volume 2: Short Papers)}, pages 567--573.

\bibitem[{Ghosh and Veale(2017)}]{ghosh2017magnets}
Aniruddha Ghosh and Tony Veale. 2017.
\newblock \href {https://www.aclweb.org/anthology/D17-1050.pdf} {Magnets for
  sarcasm: Making sarcasm detection timely, contextual and very personal}.
\newblock In \emph{Proceedings of the 2017 Conference on Empirical Methods in
  Natural Language Processing}, pages 482--491.

\bibitem[{Graefe et~al.(2018)Graefe, Haim, Haarmann, and
  Brosius}]{graefe2018readers}
Andreas Graefe, Mario Haim, Bastian Haarmann, and Hans-Bernd Brosius. 2018.
\newblock \href {https://journals.sagepub.com/doi/abs/10.1177/1464884916641269}
  {Readers’ perception of computer-generated news: Credibility, expertise,
  and readability}.
\newblock \emph{Journalism}, 19(5):595--610.

\bibitem[{Heafield(2011)}]{heafield2011kenlm}
Kenneth Heafield. 2011.
\newblock \href {https://www.aclweb.org/anthology/W11-2123.pdf} {Kenlm: Faster
  and smaller language model queries}.
\newblock In \emph{Proceedings of the sixth workshop on statistical machine
  translation}, pages 187--197. Association for Computational Linguistics.

\bibitem[{Keskar et~al.(2019)Keskar, McCann, Varshney, Xiong, and
  Socher}]{keskar2019ctrl}
Nitish~Shirish Keskar, Bryan McCann, Lav~R Varshney, Caiming Xiong, and Richard
  Socher. 2019.
\newblock \href {https://arxiv.org/pdf/1909.05858.pdf} {Ctrl: A conditional
  transformer language model for controllable generation}.
\newblock \emph{arXiv preprint arXiv:1909.05858}.

\bibitem[{Kim(2014)}]{kim2014convolutional}
Yoon Kim. 2014.
\newblock \href {https://arxiv.org/pdf/1408.5882.pdf} {Convolutional neural
  networks for sentence classification}.
\newblock In \emph{Proceedings of the 2014 Conference on Empirical Methods in
  Natural Language Processing (EMNLP)}, pages 1746--1751.

\bibitem[{Kobayashi(2018)}]{kobayashi2018contextual}
Sosuke Kobayashi. 2018.
\newblock \href {https://www.aclweb.org/anthology/N18-2072.pdf} {Contextual
  augmentation: Data augmentation by words with paradigmatic relations}.
\newblock In \emph{Proceedings of the 2018 Conference of the North American
  Chapter of the Association for Computational Linguistics: Human Language
  Technologies, Volume 2 (Short Papers)}, pages 452--457.

\bibitem[{Krizhevsky et~al.(2012)Krizhevsky, Sutskever, and
  Hinton}]{krizhevsky2012imagenet}
Alex Krizhevsky, Ilya Sutskever, and Geoffrey~E Hinton. 2012.
\newblock \href
  {http://papers.nips.cc/paper/4824-imagenet-classification-with-deep-convolutional-neural-networ}
  {Imagenet classification with deep convolutional neural networks}.
\newblock In \emph{Advances in neural information processing systems}, pages
  1097--1105.

\bibitem[{Lewis et~al.(2019)Lewis, Liu, Goyal, Ghazvininejad, Mohamed, Levy,
  Stoyanov, and Zettlemoyer}]{lewis2019bart}
Mike Lewis, Yinhan Liu, Naman Goyal, Marjan Ghazvininejad, Abdelrahman Mohamed,
  Omer Levy, Ves Stoyanov, and Luke Zettlemoyer. 2019.
\newblock \href {https://arxiv.org/pdf/1910.13461.pdf} {Bart: Denoising
  sequence-to-sequence pre-training for natural language generation,
  translation, and comprehension}.
\newblock \emph{arXiv preprint arXiv:1910.13461}.

\bibitem[{Maaten and Hinton(2008)}]{maaten2008visualizing}
Laurens van~der Maaten and Geoffrey Hinton. 2008.
\newblock \href
  {http://www.jmlr.org/papers/volume9/vandermaaten08a/vandermaaten08a.pdf}
  {Visualizing data using t-sne}.
\newblock \emph{Journal of machine learning research}, 9(Nov):2579--2605.

\bibitem[{Mikolov et~al.(2013)Mikolov, Sutskever, Chen, Corrado, and
  Dean}]{mikolov2013distributed}
Tomas Mikolov, Ilya Sutskever, Kai Chen, Greg~S Corrado, and Jeff Dean. 2013.
\newblock \href
  {https://papers.nips.cc/paper/5021-distributed-representations-of-words-and-phrases-and-their-compositionality.pdf}
  {Distributed representations of words and phrases and their
  compositionality}.
\newblock In \emph{Advances in neural information processing systems}, pages
  3111--3119.

\bibitem[{Mohler et~al.(2016)Mohler, Brunson, Rink, and
  Tomlinson}]{mohler2016introducing}
Michael Mohler, Mary Brunson, Bryan Rink, and Marc Tomlinson. 2016.
\newblock \href {https://www.aclweb.org/anthology/L16-1668.pdf} {Introducing
  the lcc metaphor datasets}.
\newblock In \emph{Proceedings of the Tenth International Conference on
  Language Resources and Evaluation (LREC'16)}, pages 4221--4227.

\bibitem[{Munos et~al.(2016)Munos, Stepleton, Harutyunyan, and
  Bellemare}]{munos2016safe}
R{\'e}mi Munos, Tom Stepleton, Anna Harutyunyan, and Marc Bellemare. 2016.
\newblock \href
  {http://papers.nips.cc/paper/6538-safe-and-efficient-off-policy-reinforcement-learning.pdf}
  {Safe and efficient off-policy reinforcement learning}.
\newblock In \emph{Advances in Neural Information Processing Systems}, pages
  1054--1062.

\bibitem[{Niu and Bansal(2018)}]{niu2018adversarial}
Tong Niu and Mohit Bansal. 2018.
\newblock \href {https://www.aclweb.org/anthology/K18-1047.pdf} {Adversarial
  over-sensitivity and over-stability strategies for dialogue models}.
\newblock In \emph{Proceedings of the 22nd Conference on Computational Natural
  Language Learning}, pages 486--496.

\bibitem[{Radford et~al.(2019)Radford, Wu, Child, Luan, Amodei, and
  Sutskever}]{radford2019language}
Alec Radford, Jeffrey Wu, Rewon Child, David Luan, Dario Amodei, and Ilya
  Sutskever. 2019.
\newblock \href
  {https://cdn.openai.com/better-language-models/language_models_are_unsupervised_multitask_learners.pdf}
  {Language models are unsupervised multitask learners}.
\newblock \emph{OpenAI Blog}, 1(8):9.

\bibitem[{Raffel et~al.(2019)Raffel, Shazeer, Roberts, Lee, Narang, Matena,
  Zhou, Li, and Liu}]{raffel2019exploring}
Colin Raffel, Noam Shazeer, Adam Roberts, Katherine Lee, Sharan Narang, Michael
  Matena, Yanqi Zhou, Wei Li, and Peter~J Liu. 2019.
\newblock \href {https://arxiv.org/pdf/1910.10683.pdf} {Exploring the limits of
  transfer learning with a unified text-to-text transformer}.
\newblock \emph{arXiv preprint arXiv:1910.10683}.

\bibitem[{Rao and Tetreault(2018)}]{rao2018dear}
Sudha Rao and Joel Tetreault. 2018.
\newblock \href {https://www.aclweb.org/anthology/N18-1012.pdf} {Dear sir or
  madam, may i introduce the gyafc dataset: Corpus, benchmarks and metrics for
  formality style transfer}.
\newblock In \emph{Proceedings of the 2018 Conference of the North American
  Chapter of the Association for Computational Linguistics: Human Language
  Technologies, Volume 1 (Long Papers)}, pages 129--140.

\bibitem[{Reimers and Gurevych(2019)}]{reimers2019sentence}
Nils Reimers and Iryna Gurevych. 2019.
\newblock \href {https://arxiv.org/pdf/1908.10084.pdf} {Sentence-bert: Sentence
  embeddings using siamese bert-networks}.
\newblock In \emph{Proceedings of the 2019 Conference on Empirical Methods in
  Natural Language Processing and the 9th International Joint Conference on
  Natural Language Processing (EMNLP-IJCNLP)}, pages 3973--3983.

\bibitem[{Schulman et~al.(2017)Schulman, Wolski, Dhariwal, Radford, and
  Klimov}]{schulman2017proximal}
John Schulman, Filip Wolski, Prafulla Dhariwal, Alec Radford, and Oleg Klimov.
  2017.
\newblock \href {https://arxiv.org/pdf/1707.06347.pdf} {Proximal policy
  optimization algorithms}.
\newblock \emph{arXiv preprint arXiv:1707.06347}.

\bibitem[{Sennrich et~al.(2016)Sennrich, Haddow, and
  Birch}]{sennrich2016improving}
Rico Sennrich, Barry Haddow, and Alexandra Birch. 2016.
\newblock \href {https://www.aclweb.org/anthology/P16-1009.pdf} {Improving
  neural machine translation models with monolingual data}.
\newblock In \emph{Proceedings of the 54th Annual Meeting of the Association
  for Computational Linguistics (Volume 1: Long Papers)}, pages 86--96.

\bibitem[{Severyn and Moschitti(2015)}]{severyn2015twitter}
Aliaksei Severyn and Alessandro Moschitti. 2015.
\newblock \href {https://dl.acm.org/doi/pdf/10.1145/2766462.2767830} {Twitter
  sentiment analysis with deep convolutional neural networks}.
\newblock In \emph{Proceedings of the 38th International ACM SIGIR Conference
  on Research and Development in Information Retrieval}, pages 959--962.

\bibitem[{Silfverberg et~al.(2017)Silfverberg, Wiemerslage, Liu, and
  Mao}]{silfverberg2017data}
Miikka Silfverberg, Adam Wiemerslage, Ling Liu, and Lingshuang~Jack Mao. 2017.
\newblock \href {https://www.aclweb.org/anthology/K17-2010.pdf} {Data
  augmentation for morphological reinflection}.
\newblock In \emph{Proceedings of the CoNLL SIGMORPHON 2017 Shared Task:
  Universal Morphological Reinflection}, pages 90--99.

\bibitem[{Szegedy et~al.(2015)Szegedy, Liu, Jia, Sermanet, Reed, Anguelov,
  Erhan, Vanhoucke, and Rabinovich}]{szegedy2015going}
Christian Szegedy, Wei Liu, Yangqing Jia, Pierre Sermanet, Scott Reed, Dragomir
  Anguelov, Dumitru Erhan, Vincent Vanhoucke, and Andrew Rabinovich. 2015.
\newblock \href
  {https://www.cv-foundation.org/openaccess/content_cvpr_2015/html/Szegedy_Going_Deeper_With_2015_CVPR_paper.html}
  {Going deeper with convolutions}.
\newblock In \emph{Proceedings of the IEEE Conference on Computer Vision and
  Pattern Recognition}, pages 1--9.

\bibitem[{Vaswani et~al.(2017)Vaswani, Shazeer, Parmar, Uszkoreit, Jones,
  Gomez, Kaiser, and Polosukhin}]{vaswani2017attention}
Ashish Vaswani, Noam Shazeer, Niki Parmar, Jakob Uszkoreit, Llion Jones,
  Aidan~N Gomez, {\L}ukasz Kaiser, and Illia Polosukhin. 2017.
\newblock \href
  {http://papers.nips.cc/paper/7181-attention-is-all-you-need.pdf} {Attention
  is all you need}.
\newblock In \emph{Advances in neural information processing systems}, pages
  5998--6008.

\bibitem[{Wang and Yang(2015)}]{wang_yang_2015_thats}
William~Yang Wang and Diyi Yang. 2015.
\newblock \href {https://doi.org/10.18653/v1/D15-1306} {That{'}s so
  annoying!!!: A lexical and frame-semantic embedding based data augmentation
  approach to automatic categorization of annoying behaviors using {\#}petpeeve
  tweets}.
\newblock In \emph{Proceedings of the 2015 Conference on Empirical Methods in
  Natural Language Processing}, pages 2557--2563, Lisbon, Portugal. Association
  for Computational Linguistics.

\bibitem[{Waseem(2016)}]{waseem:2016:NLPandCSS}
Zeerak Waseem. 2016.
\newblock \href {http://aclweb.org/anthology/W16-5618} {Are you a racist or am
  i seeing things? annotator influence on hate speech detection on twitter}.
\newblock In \emph{Proceedings of the First Workshop on NLP and Computational
  Social Science}, pages 138--142, Austin, Texas. Association for Computational
  Linguistics.

\bibitem[{Wei and Zou(2019)}]{wei2019eda}
Jason Wei and Kai Zou. 2019.
\newblock \href {https://www.aclweb.org/anthology/D19-1670.pdf} {Eda: Easy data
  augmentation techniques for boosting performance on text classification
  tasks}.
\newblock In \emph{Proceedings of the 2019 Conference on Empirical Methods in
  Natural Language Processing and the 9th International Joint Conference on
  Natural Language Processing (EMNLP-IJCNLP)}, pages 6383--6389.

\bibitem[{Xie et~al.(2017)Xie, Wang, Li, L{\'e}vy, Nie, Jurafsky, and
  Ng}]{xie2017data}
Ziang Xie, Sida~I Wang, Jiwei Li, Daniel L{\'e}vy, Aiming Nie, Dan Jurafsky,
  and Andrew~Y Ng. 2017.
\newblock \href {https://arxiv.org/pdf/1703.02573.pdf} {Data noising as
  smoothing in neural network language models}.
\newblock \emph{arXiv preprint arXiv:1703.02573}.

\bibitem[{Yang et~al.(2020)Yang, Malaviya, Fernandez, Swayamdipta, Le~Bras,
  Wang, Bhagavatula, Choi, and Downey}]{yang2020g}
Yiben Yang, Chaitanya Malaviya, Jared Fernandez, Swabha Swayamdipta, Ronan
  Le~Bras, Ji-Ping Wang, Chandra Bhagavatula, Yejin Choi, and Doug Downey.
  2020.
\newblock \href {https://arxiv.org/pdf/2004.11546.pdf} {G-daug: Generative data
  augmentation for commonsense reasoning}.
\newblock \emph{arXiv}, pages arXiv--2004.

\bibitem[{Yang et~al.(2019)Yang, Dai, Yang, Carbonell, Salakhutdinov, and
  Le}]{yang2019xlnet}
Zhilin Yang, Zihang Dai, Yiming Yang, Jaime Carbonell, Russ~R Salakhutdinov,
  and Quoc~V Le. 2019.
\newblock \href
  {http://papers.nips.cc/paper/8812-xlnet-generalized-autoregressive-pretraining-for-language-understanding.pdf}
  {Xlnet: Generalized autoregressive pretraining for language understanding}.
\newblock In \emph{Advances in neural information processing systems}, pages
  5754--5764.

\bibitem[{Yu et~al.(2018)Yu, Dohan, Le, Luong, Zhao, and Chen}]{yu2018qanet}
Adams~Wei Yu, David Dohan, Quoc Le, Thang Luong, Rui Zhao, and Kai Chen. 2018.
\newblock \href {https://openreview.net/forum?id=B14TlG-RW} {Fast and accurate
  reading comprehension by combining self-attention and convolution}.
\newblock In \emph{International Conference on Learning Representations}.

\bibitem[{Zhang et~al.(2015)Zhang, Zhao, and LeCun}]{zhang2015character}
Xiang Zhang, Junbo Zhao, and Yann LeCun. 2015.
\newblock \href
  {http://papers.nips.cc/paper/5782-character-level-convolutional-networks-for-text-classification.pdf}
  {Character-level convolutional networks for text classification}.
\newblock In \emph{Advances in neural information processing systems}, pages
  649--657.

\bibitem[{Zhou et~al.(2016)Zhou, Shi, Tian, Qi, Li, Hao, and
  Xu}]{zhou2016attention}
Peng Zhou, Wei Shi, Jun Tian, Zhenyu Qi, Bingchen Li, Hongwei Hao, and Bo~Xu.
  2016.
\newblock \href {https://www.aclweb.org/anthology/P16-2034.pdf}
  {Attention-based bidirectional long short-term memory networks for relation
  classification}.
\newblock In \emph{Proceedings of the 54th annual meeting of the association
  for computational linguistics (volume 2: Short papers)}, pages 207--212.

\end{thebibliography}

\end{document}